\documentclass{article}

\usepackage{arxiv}

\usepackage[utf8]{inputenc} % allow utf-8 input
\usepackage[T1]{fontenc}    % use 8-bit T1 fonts
\usepackage{hyperref}       % hyperlinks
\usepackage{url}            % simple URL typesetting
\usepackage{booktabs}       % professional-quality tables
\usepackage{amsfonts}       % blackboard math symbols
\usepackage{nicefrac}       % compact symbols for 1/2, etc.
\usepackage{microtype}      % microtypography
\usepackage{lipsum}		% Can be removed after putting your text content
\usepackage{graphicx}
\usepackage{natbib}
\usepackage{doi}
\usepackage{multirow}
\usepackage{caption,subcaption}

\title{DeepDamageNet: A two-step deep-learning model for multi-disaster building damage segmentation and classification using satellite imagery}

\author{Irene Alisjahbana \qquad Jiawei Li \qquad Ben (Mullet) Strong \qquad Yue Zhang \\
Stanford University\\
{\tt\small alisjahbana@stanford.edu,
jiaweili@stanford.edu,}\\ 
{\tt\small ben.gm.strong@gmail.com, yzhang16@stanford.edu}
}

\renewcommand{\headeright}{}
\renewcommand{\shorttitle}{}

\begin{document}
\maketitle

\begin{abstract}
    Satellite imagery has played an increasingly important role in post-disaster building damage assessment. Unfortunately, current methods still rely on manual visual interpretation, which is often time-consuming and can cause very low accuracy. To address the limitations of manual interpretation, there has been a significant increase in efforts to automate the process. We present a solution that performs the two most important tasks in building damage assessment, segmentation and classification, through deep-learning models. We show our results submitted as part of the xView2 Challenge, a competition to design better models for identifying buildings and their damage level after exposure to multiple kinds of natural disasters. Our best model couples a building identification semantic segmentation convolutional neural network (CNN) to a building damage classification CNN, with a combined F1 score of 0.66, surpassing the xView2 challenge baseline F1 score of 0.28. We find that though our model was able to identify buildings with relatively high accuracy, building damage classification across various disaster types is a difficult task due to the visual similarity between different damage levels and different damage distribution between disaster types, highlighting the fact that it may be important to have a probabilistic prior estimate regarding disaster damage in order to obtain accurate predictions. 
\end{abstract}

% keywords can be removed
\section{Introduction}

In the past two decades, the number of natural disasters that occur worldwide has increased dramatically, causing thousands of casualties and millions of dollars of loss to buildings every year. It is no surprise that rapid building damage assessment is considered to be one of the most important phases after major disasters, aiming to evaluate the state of the buildings after a given event. In addition to providing information to first responders for rescue operations and relief efforts, it also provides useful information for local government and related organizations for the estimation of economic loss and damage distribution in the subsequent reconstruction and recovery phase \cite{barrington2012crowdsourcing}. Though significant efforts have been put into rapid post-disaster assessment, it remains a very challenging task.

Traditional methods of building damage assessments rely on ground-based surveys and field assessments to assess building damage, which involves dispatching experts to the field to manually inspect buildings individually. Though it is still perhaps the most reliable method of assessment, this method is often time-consuming to apply across large geographic regions, subjective and prone to inconsistencies in data quality \cite{brown2015monitoring}. 

Technological advancements in the remote sensing field have created new opportunities to address the limitations of field-based assessments and have been recognized to be useful for rapid post-disaster response \cite{rathje2008role, dong2013comprehensive, taubenbock2014capabilities, ghaffarian2018remote}. It enables the collection of data that is independent and rapidly available across large geographical regions. Unfortunately, most efforts for remote sensing-based building damage assessment today still rely on manual visual interpretation and volunteers, which can result in very low accuracy \cite{Loos2018TheRemote-Sensing}. Though existing emergency mapping services such as the Copernicus Emergency Management Service (EMS) provide damage maps a few hours after a disaster, they are still typically carried out by human operators \cite{Nex2019TowardsSolutions}. This involves manually detecting and segmenting the buildings before classifying them based on their damage level. 

To address the limitations of manual interpretation, there has been a significant increase in efforts to automate the process. Research has shown that using machine learning and deep learning techniques, in particular convolution neural networks (CNN) can achieve high accuracy. In addition, the use of automated methods allows the independent and rapid assessment that can easily be applied across large geographical areas. Nonetheless, most of the work currently being done focuses on the two tasks (segmentation and classification) separately. 

To that end, the xView2 Challenge aims to automate the process of building damage assessment from pre- and post-disaster satellite imagery through the creation of accurate and efficient machine learning models. To support the advancement of the automation process, the xBD dataset \cite{gupta2019creating}, a new large-scale dataset containing pre- and post- disaster satellite imagery and ground truth labels from a variety of disasters and over 160,000 buildings was released. 

Through this challenge, we present a framework to perform segmentation and classification of buildings from pairs of pre- and post- disaster images for the purposes of building damage assessment. In particular, we explored various methods and model architectures that would be generalizable across various types of natural disasters, such as earthquakes, tsunamis, floods, volcanic eruptions, wildfires, and hurricanes. We present our results and analyses on the advantages and limitations of using automated methods for the purposes of building damage assessment across different disasters. 

\section{Related Work}
Given its obvious advantages, damage assessment via remote sensing has become a topic of much research in the last few decades. Verstappen \cite{verstappen1995aerospace} proposed one of the original frameworks for evaluating the use of remote sensing data in natural hazard risk reduction, and called for the use of remote sensing tools in combating both “instantaneous” disasters like cyclones, drought, floods, tsunamis, earthquakes, and volcanic hazards, and “creeping” disasters like desertification. Frameworks have also been proposed for the use of remote sensing data to predict high-risk areas, before disaster strikes \cite{tralli2005satellite}.

In the pre-machine learning era, practitioners relied on classical segmentation and classification techniques for map production, e.g.  a “region-growing and knowledge-based segmentation approach” that involves merging visually similar regions to build a disaster map \cite{van2003segmentation}. It is only more recently that researchers have applied machine learning techniques to this problem, with increasing degrees of success. 

Thanks to the recent advancement of the object detection and segmentation algorithms in computer vision, many CNN models have been developed and applied for building footprint segmentation based on satellite data. Some of the most common models include U-Net \cite{ronneberger2015u}, and the more recent feature pyramid network (FPN) \cite{lin2017feature}. U-Net has been successfully used in the satellite image building segmentation challenge SpaceNet 4. Chen et al. \cite{chen2018aerial} used the FPN model for building segmentation using the Aerial Imagery for Roof Segmentation (AIRS) dataset. This dataset contains aerial imagery with 7.5cm resolution and contains over 220,000 buildings. The overall performance of the FPN model was evaluated through intersection over union (IoU) and F1-score metrics, in which they achieved a score of 0.882 and 0.937 respectively. 

One limitation of using semantic segmentation for the purposes of building damage assessment is the fact that semantic segmentation treats multiple objects of the same class as a single entity, whereas in reality, each building is required to be classified individually. As a result, instance segmentation has been increasingly used, as it identifies multiple objects of the same class as distinct individual objects \cite{chen2018masklab, iglovikov2018ternausnetv2}.  He et al. \cite{he2017mask} first introduced Mask R-CNN, a framework for generating high-quality segmentation masks for each instance. Since then, Mask R-CNN has been used for a variety of different application as it is fast to implement and easy to generalize to different tasks including building footprint segmentation. One such application was for CrowdAI's mapping challenge 2018 \cite{crowdAIMappingChallengeBaseline2018}, aiming to segment buildings for humanitarian assistance. Using high resolution RGB satellite images, the challenge baseline achieved an IoU of 0.697.

In addition to building footprint segmentation, deep learning methods are widely used to classify the building damage level. Xu et al. \cite{xu2019building} compared four architectures to perform binary classification (damage or not damaged) for a certain building cropped from a satellite imagery. Their results show that a twin-tower structure to fuse the pre-disaster and post-disaster images can improve the classification accuracy. In their experiments, the best model used two separate CNNs to extract features of the pre- and post-disaster images, before subtracting the extracted features along the
channel dimension. They used the area under the RoC curve (AUC) as the primary metric of model, and achieved a score of 0.8302 AUC score on the Haiti earthquake dataset.

As seen above, most work focuses on either segmentation or classification. Gupta et al. \cite{gupta2019creating} introduced a baseline model for the xView2 challenge that combines the segmentation and classification model. The segmentation model is based on an altered U-Net architecture, achieving an IoU of 0.66 and F1 score of 0.80. On the other hand, the classification model is based on a modified ResNet50. The weighted F1 is used as the primary metric for evaluating the overall dataset, obtaining a weighted F1 score of 0.2564. 

To our knowledge, no previous work has been done to create a model that would be generalizable to all types of natural disasters. Most work have focused on a single disaster type. For example, Xu et al. \cite{xu2019building} trained a model for earthquake disasters damage whereas Rudner et al. \cite{rudner2019multi3net} for flooded buildings. Xu et al. \cite{xu2019building} quantified how well the models will generalize to future disasters by training and testing models on different disaster events, and found that the AUC reduces for cross-region results. This suggests that cross-region prediction is difficult, even within the same disaster type (i.e earthquakes). 

\section{Dataset}
\label{dataset}

\subsection{Dataset Description}
\begin{figure}[h!]
\begin{center}
    
    \includegraphics[width=5in]{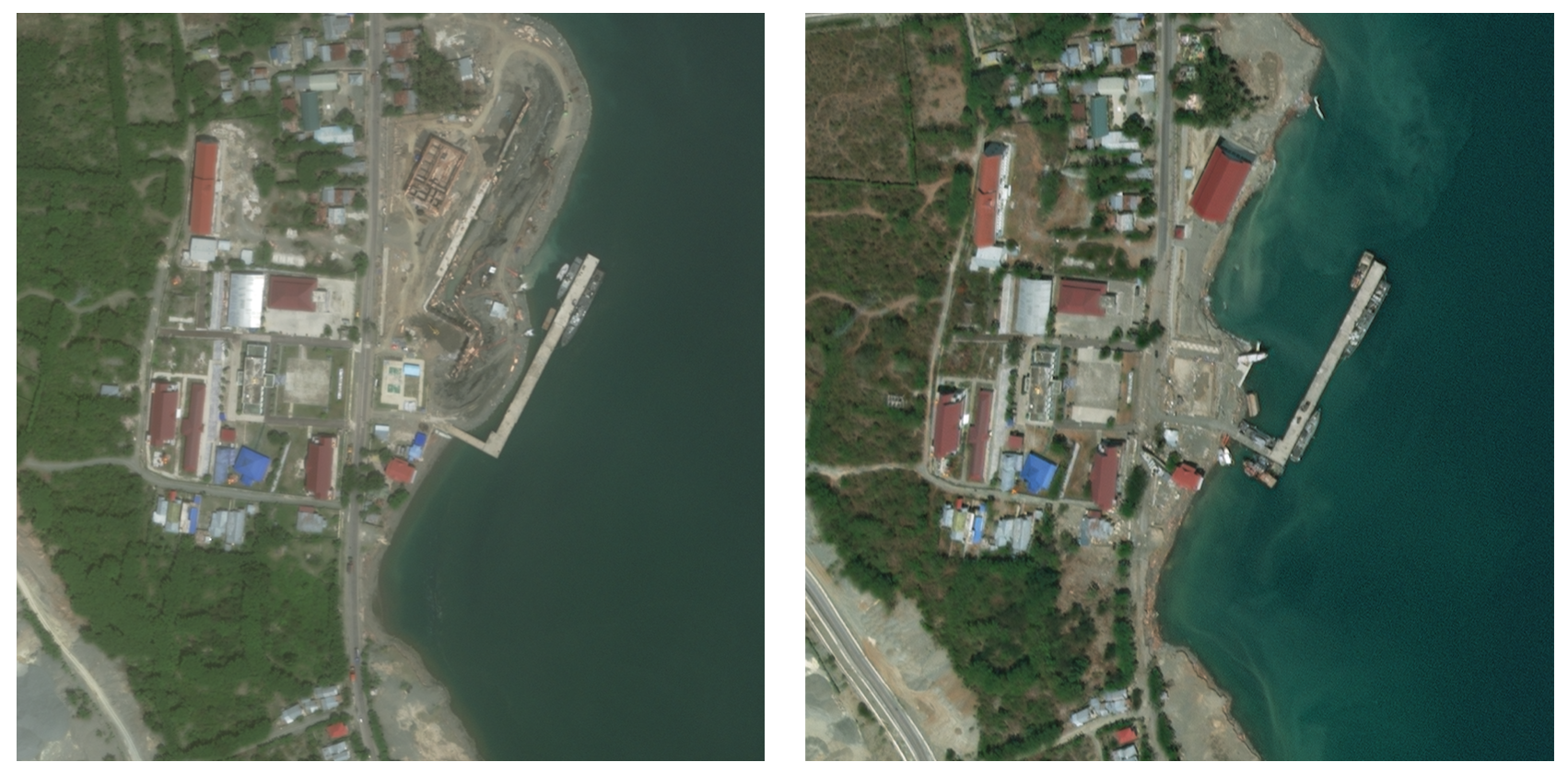}
\end{center}
    \caption{Pre-disaster (left) and its corresponding post-disaster satellite image }
\label{fig:prepostimage}
\end{figure}

The dataset used is the xBD dataset, released to supplement the xView2 Challenge \cite{gupta2019creating}. A total of 2799 pairs of pre- and post-disaster RGB satellite images are provided in the training set, whereas  933 pairs of pre- and post- disaster satellite images are available in the test set. All imagery is obtained from DigitalGlobe satellite with a resolution of around 0.5m ground sample distance. The size of the images are $1024 \times 1024$ pixels. An example of a pre- and post- disaster image can be seen in Figure \ref{fig:prepostimage}.

In addition to imagery, the xBD dataset provides ground truth data that includes the coordinates of each building polygons and the corresponding damage scale level for each building. The damage scale level that is used in the xBD dataset is a joint damage scale that ranges from no damage (0) to destroyed (3). The ground truth data is only available for the training set. A detailed description of the damage scale definition, image collection method, annotation process and quality control of the dataset is available in \cite{gupta2019creating}.

\subsection{Dataset Statistics}

 The training dataset consists of 10 different disasters, ranging from hurricanes, to earthquakes and is gathered from over 15 countries at various times of the year. It should be noted that the number of images available for each disaster event is highly unbalanced, with the Guatemala volcanic eruption having the least number of images.

\begin{figure}[h!]
\begin{center}
    \includegraphics[width=5in]{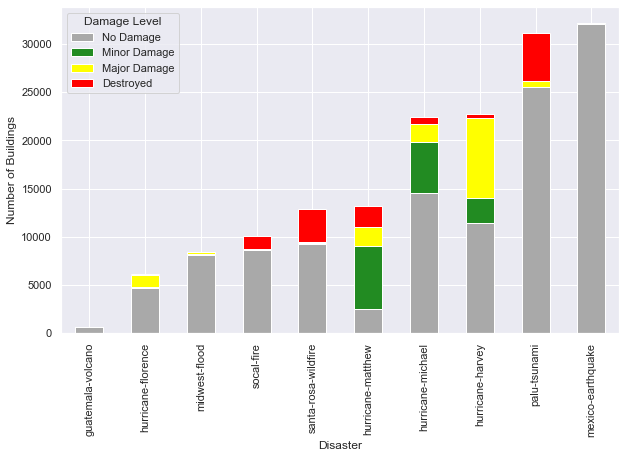}
\end{center}
  \caption{Distribution of building damage level in training set}
\label{fig:building_dist}
\end{figure}

\begin{figure}[htp]
\begin{center}
    \includegraphics[width=5in]{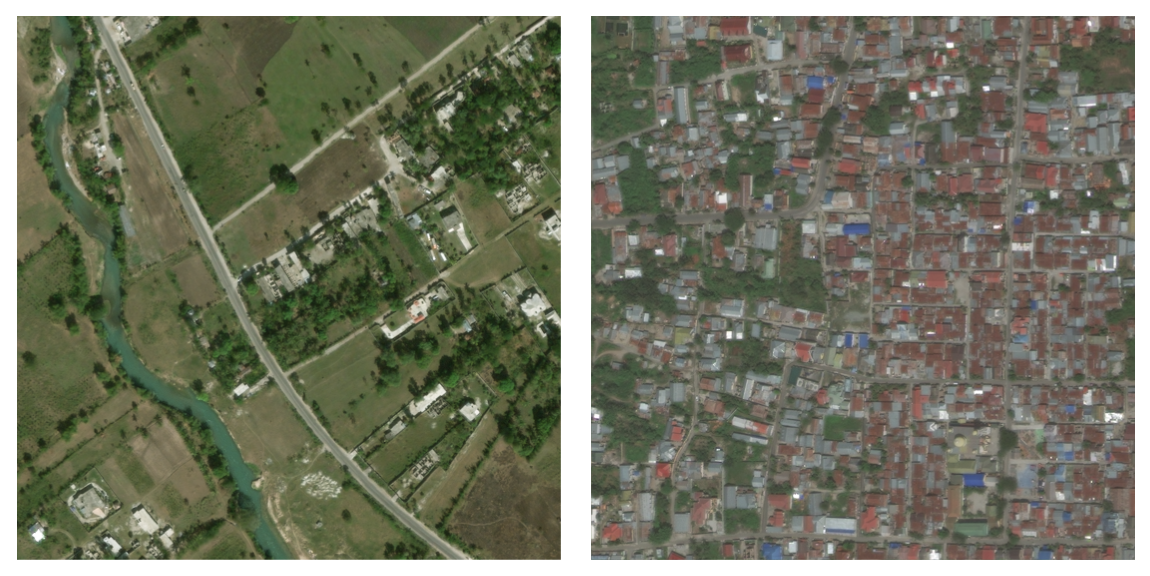}
\end{center}
   \caption{Differences in building densities}
\label{fig:building_density}
\end{figure}

Though the distribution of images based on each disaster is highly imbalanced, the number of buildings that are present in each image also varies significantly between each disaster (Figure \ref{fig:building_dist}). For some disasters, there are images where a single image contains over 1000 buildings. On the other hand, there are disasters with images that contain less than 10 buildings (Figure \ref{fig:building_density}). 

Furthermore, the damage level for the buildings in the training dataset is also imbalanced. Non-damaged building takes up over 70\% of the training data. However, even though there are mostly non-damage buildings in the complete training data, this distribution varies significantly between each disaster as seen in Figure \ref{fig:building_density}. For example, there are more damaged buildings compared to non-damaged buildings for Hurricane Matthew. 
% The various imbalances discussed previously are all taken into account when creating the training/validation split. 

\section{Methods}
\subsection{Pre-Processing}
Before splitting the xBD dataset into train/val set, we manually looked through the dataset for features that might be problematic for our model. For example, there were some images that are covered by clouds, causing occlusion to some buildings. However, the statistical analysis showed that less than 1\% of the images were covered by clouds. As a result, we decided to leave those images in our dataset.  

% The other issue is the padding values. Even though all input images are of the same size, some images are padded with value 0, which should not be considered in the model development. We labeled those pixels as 127 and ignored them in training and evaluation. 

We then split the data into training and validation with 80\% to 20\% ratio. Since the hold-out test set used by the challenge is drawn from the same disaster type distribution as the data released, our validation set was created by preserving this distribution. For each disaster, we took 20\% of the images for the validation set, and 80\% for the training set. The number of images in the training and validation set for each disaster can be seen in Table \ref{tab:split}.

\begin{table}
\begin{center}
\begin{tabular}{|l|c|c|c|}
\hline
Disaster & Total & Train. Set & Val. Set \\
\hline\hline
Guatemala Volcano & 18 & 15 & 3 \\
Hurricane Florence & 319 & 256 & 63 \\
Hurricane Harvey & 319 & 256 & 63 \\
Hurricane Matthew & 238 & 191 & 47 \\
Hurricane Michael & 343 & 191 & 68 \\
Mexico Earthquuake & 121 & 97 & 24 \\
Midwest Flood & 279 & 224 & 55 \\
Palu Tsunami & 113 & 91 & 22 \\
Santa Rosa Wildfire & 226 & 181 & 45 \\
Socal Fire & 823 & 659 & 164 \\
\hline\hline
Total Images & 2799 & 2245 & 554 \\
\hline
\end{tabular}
\end{center}
\caption{Number of images in the training and validation sets based on disaster}
\label{tab:split}
\end{table}

We recognize that even with the split generated above, the building damage class imbalance persists. A majority of the buildings in our split dataset will still be non-damaged buildings. In our current approach, we did not do any further distribution of data in order to reduce this imbalance.

% However, one issue with this method was that the damage level distributions for each sampled image are very different. Some images have almost all buildings with no-damage and one or two with major damage. Other images may have all buildings destroyed. There is a chance that in one certain split, there are too many buildings of one damage level in the validation set and too few in the training set. The model then cannot learn how to classify that level correctly. Thus, for the split we generated above, we checked the building damage level distribution for both training and validation and make sure we have similar distribution for each damage level.

\subsection{Model Architecture}
\begin{figure}[h!]
\begin{center}
    \includegraphics[width=6.5in]{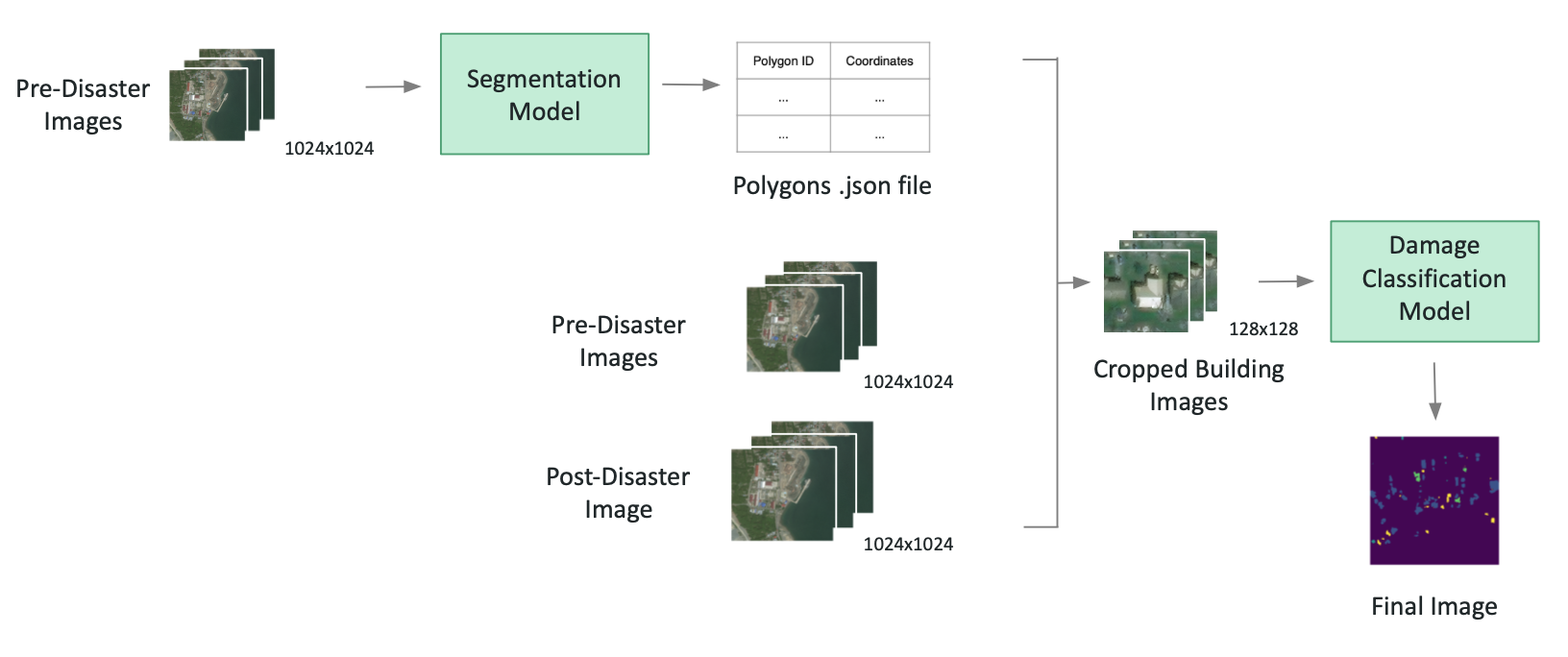}
\end{center}
  \caption{Architecture of two-step model. The pre-disaster image is fed into a segmentation model to obtain building polygon coordinates. This result in addition to pre- and post- disaster cropped building image is then fed into a damage classification model to obtain the final image.}
\label{fig:twostep}
\end{figure}

We experimented with two techniques for performing the objective task. First, we built and trained an end-to-end model, which was designed to simultaneously perform both segmentation of building footprints and classification of damage extent. Second, we implemented a series of two-step models, which treated building segmentation and damage classification as two distinct tasks, with separate models optimized for each of these tasks (Figure \ref{fig:twostep}).

\subsubsection{End-to-End Model}
\begin{figure}[h!]
\begin{center}
    \includegraphics[width=5.5in]{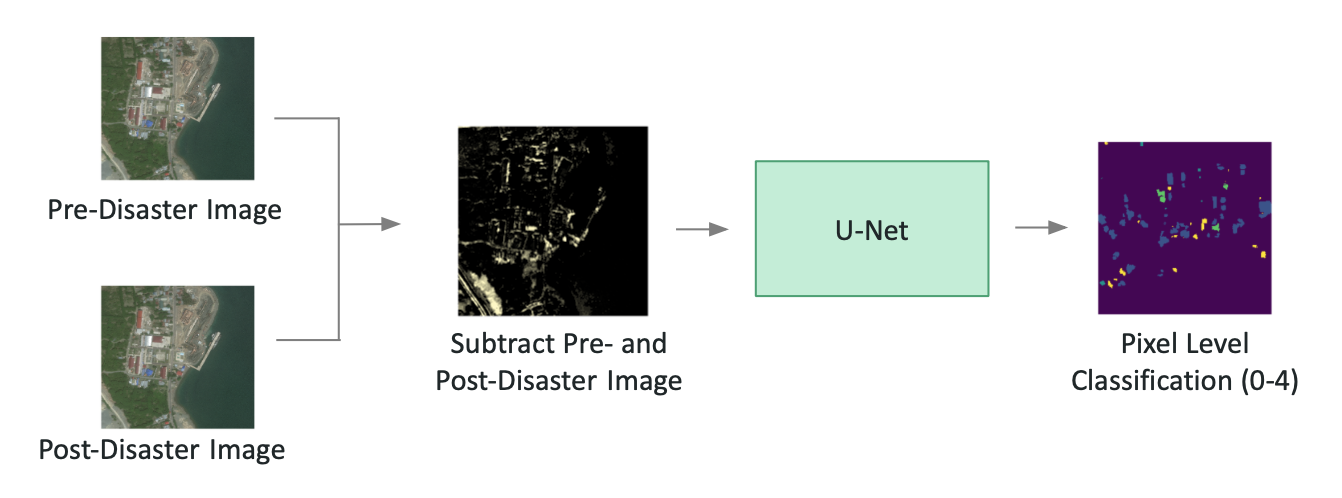}
\end{center}
  \caption{Architecture of end-to-end model. The pre- and disaster image is subtracted before feeding into a U-Net to perform simultaneous segmentation and classification.}
\label{fig:endtoend}
\end{figure}

For the end-to-end model, the inputs were paired pre-disaster and post-disaster images both of size $1024 \times 1024 \times 3$ and the output shows the pixel-level classification results of size $1024 \times 1024$. The classification values ranged within 0 to 4, where 0 represents no-building and 1-4 represented different building damage levels from no-damage to destroyed, as discussed above. The model architecture chosen here was U-Net with a ResNet-34 encoder pre-trained on ImageNet dataset. 
%Research has shown that using a pre-trained encoder yields better results than training everything from scratch \cite{thomas_2019}.
To incorporate both pre-disaster image and post-disaster image, two architectures were proposed (Figure \ref{fig:endtoend}). 

The intuitive way to make use of the pre- and post- disaster images is to compare the difference between them. Thus, we subtracted the pre-disaster image pixel values from the post-disaster image to obtain a new image of size $1024 \times 1024 \times 3$. This was then treated as a new input to be fed directly to the ResNet without making any further changes. 

%The other architecture we experimented on was to concatenate the pre-disaster and post-disaster images to get a 6-channel input. For this case, the pre-trained weights in ResNet-34 was modified to accommodate the input. The kernel size for the first layer of ResNet-34 was changed from 3 to 6 and we reinitialized the 6 kernels with the mean value of the original 3 kernel weights. 

\subsubsection{Two-Step Model: Building Segmentation}
\label{segmentation_method}

In the two-step model, the first step is to segment the building footprint and extract the polygon coordinates of each building. We tried two different model architectures for this task, semantic segmentation and instance segmentation, which are commonly used for building footprint segmentation. 

\textbf{Instance Segmentation:} The objective of the instance segmentation is to label each foreground pixel with a corresponding object and instance - a combination of semantic segmentation and object detection. For this, we used Mask R-CNN~\cite{he2017mask}. For ease of implementation, we followed the implementation from the Facebook AI research platform Detectron2~\cite{wu2019detectron2}.
The backbone model of the Mask R-CNN is a ResNet-50~\cite{he2016deep} with Feature Pyramid Network (FPN)~\cite{lin2017feature}. A faster R-CNN~\cite{ren2015faster} model is combined with the backbone model to perform the semantic segmentation and object detection. 

\textbf{Semantic Segmentation:} A ResNet-50 Feature Pyramid Network (the same backbone model as the Mask R-CNN) was used to conduct the semantic segmentation of the building footprint. The model input is the pre-disaster image and the output is the binary mask of each image with 0 (background) and 1 (building).  After obtaining the semantic mask of each image, we outlined every single building to extract their polygon coordinates using Imantics \cite{image}, a Python package. 

\subsubsection{Two step model: Building Classification}
\label{classification_model}

After segmentation maps were generated using the techniques outlined in the previous section, building-by-building damage classification was performed. The model takes as input a close cropping of an individual building in both pre- and post-disaster images and predicts the damage level score (1 for no damage, 4 for destroyed). The basic structure of the damage classification model includes two CNN feature extractors which are used to pre-process the pre- and post-disaster building images. Then the features maps can be used alone or combined with other features to conduct the prediction.

% To enable easy implementation and comparison of models, we built a damage classification model pipeline inspired by the DeepLab pipeline \cite{chen2017deeplab}.

\textbf{Pre \& Post Concatenation:} 

\begin{figure}[h!]
\begin{subfigure}{.9\textwidth}
    \includegraphics[width=\linewidth]{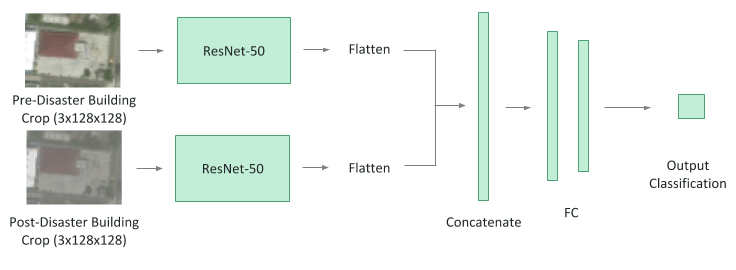}
    \caption{Pre \& Post Concatenation}
    \label{fig:classification_model_ttc}
\end{subfigure}
\newline
\begin{subfigure}{.9\textwidth}
    \includegraphics[width=\linewidth]{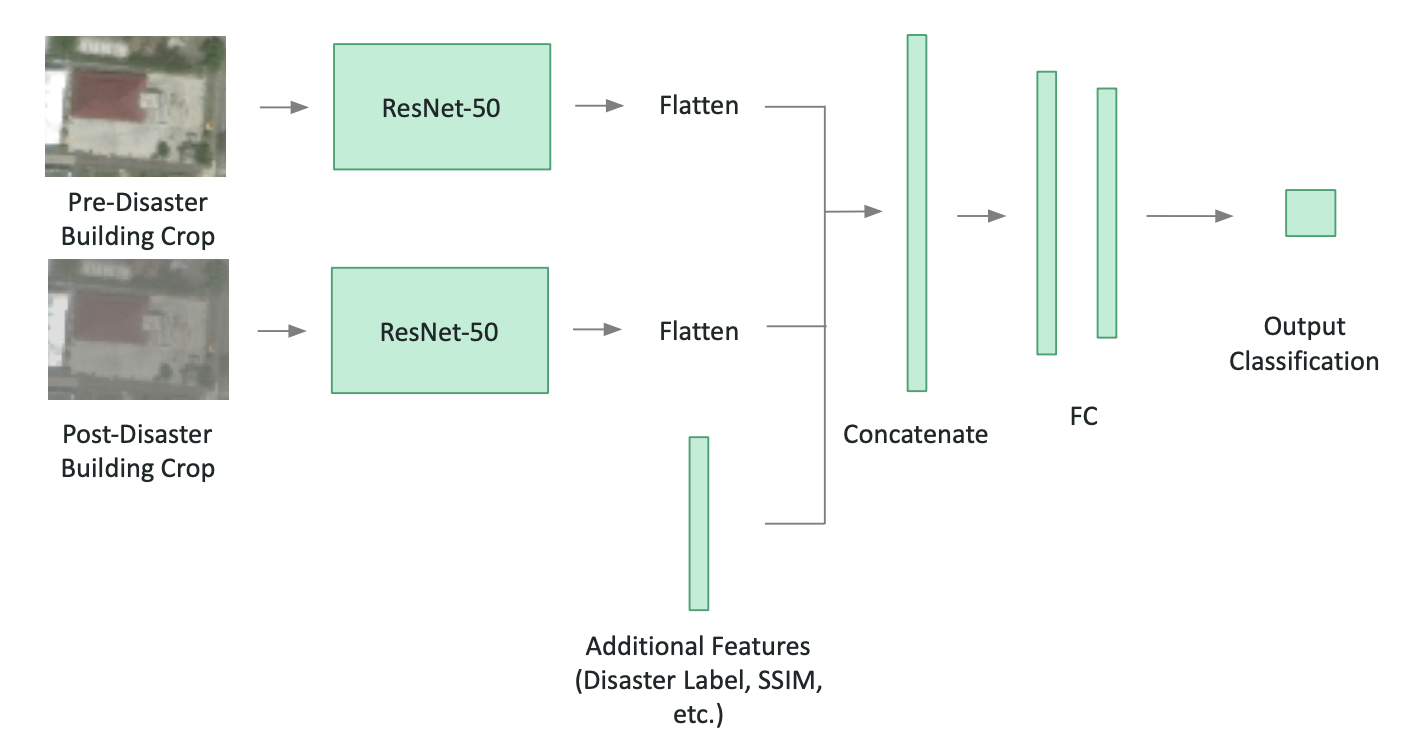}
    \caption{Pre \& Post Concatenation with Additional Features}
     \label{fig:classification_model_extra_features}
\end{subfigure}
   \caption{Architecture of classification model. a) A twin-tower architecture concatenates the output features of a ResNet-50 model from pre- and post- disaster cropped building images. b) An additional third branch can be added to include additional features.}
\label{fig:classification_model}
\end{figure}

The damage classification model concatenating pre- and post- feature maps can be seen in Figure \ref{fig:classification_model_ttc}. Similar to what was done by Xu et al. \cite{xu2019building}, we implemented a twin-tower model architecture which concatenates the output features of a ResNet-50 model (pre-trained on ImageNet). This concatenated feature map is then flattened and fed to several fully connected layers, which outputs a damage level classification score ranging from 1 (No damage) to 4 (Destroyed).

\textbf{Pre \& Post Concatenation with Disaster Prediction:} 
As mentioned before, different disasters in different regions will be covered in this single prediction model. We found that the building damage characteristics of different disasters are significantly different, which motivates us to incorporate the disaster-type as an extra features in the Pre \& Post Concatenation model in Figure \ref{fig:classification_model_extra_features}.

The xview2 dataset provides a unique disaster-type label for each training image, which enables us to train a separate disaster type prediction models. The disaster-type will then be represented as an one-hot vector and concatenated with the pre- and post- features extracted from CNN and the combined features are used for the final prediction.

\textbf{Pre \& Post Concatenation with SSIM:} 
As an extension, we also experimented by adding a third branch with additional features before concatenating and feeding to the fully connected layers. The structural similarity index (SSIM) was added as the additional feature, a metric that measures the similarity between two images \cite{wang2004image}. 

%\textbf{TODO: add more detailed description of disaster classification model}

\section{Experiments}
\subsection{Evaluation Metrics}
The evaluation metrics used to reflect the overall performance of our models are the mean Intersection-Over-Union (mIoU) and weighted F1-score. mIoU is specifically used for our segmentation model, as it is commonly used in literature to evaluate segmentation models. A weighted F1 score was also used for both the segmentation and classification models. This metric balances precision and recall in a harmonic mean and is well-suited for imbalanced datasets such as the xBD dataset \cite{gupta2019creating}. 

For the purposes of the xView2 challenge, the final evaluation metric is a combined F1 score: 30\% segmentation F1 score + 70\% classification F1 score. This metric is calculated over the hold-out test dataset. 

\subsection{Results}

\subsubsection{End-to-End Model}
\begin{figure}[h!]
\begin{center}
    \includegraphics[width=5in]{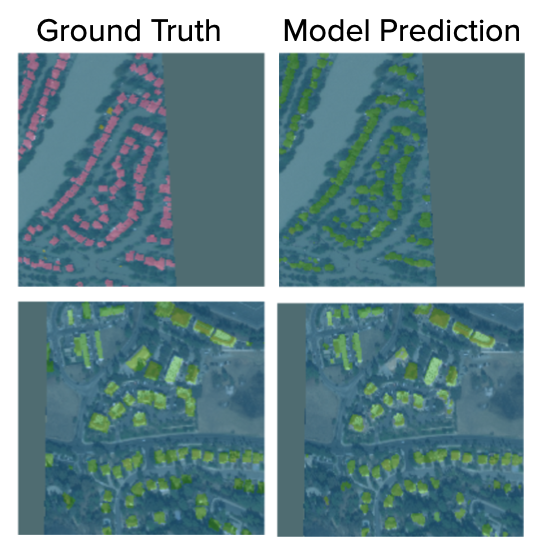}
\end{center}
   \caption{Ground truth data (left column) and model predictions (right column) for end-to-end model. Our model accurately predicts minor damage buildings (green) in some images while completely missing destroyed buildings (red) in others.}
\label{fig:endtoend_result}
\end{figure}

Examples of results obtained with our end-to-end model can be seen in Figure \ref{fig:endtoend_result}. It can be seen that our model does relatively well for the segmentation task. Our best model obtained a 0.795 F1 score on the test dataset for the segmentation task, similar to what was obtained in the xView2 baseline model. Our error analysis showed that we missed some small buildings and those buildings partially hidden in the forest/water.

% Regarding the building damage level classification, the model returned 0 F1 score on the test dataset. This is because the F1 score for the damage level classification was determined by the harmonic mean of the 4 damage levels. Thus, if the model fails at any damage level, the overall performance will be poor. 

% Furthermore, though we can assure the disaster type distribution for the test dataset images are close to our own training and validation set, the damage level distribution is unknown and can be drastically different from our training and validation set. As a result, it is hard to predict how our model performs on the test dataset. 
However, the end-to-end model was not able to accurately classify the damage level of each building and performed significantly worse than the baseline model. We realized that an end-to-end model is unable to accurately capture all the complexities of building damage levels, and therefore we proceeded to focus on the two-step model. 

\subsubsection{Building Segmentation}
As described in Section \ref{segmentation_method}, we experimented with two different segmentation architectures.

\textbf{Instance segmentation:} We trained a mask R-CNN model with two P100 GPUs using a batch size of 4 and the initial learning rate of 0.01. The original high-resolution image with dimensions $1024 \times 1024 \times 3$ was directly used as the training images. The number of regions of interest (ROIs) per image was set to be 1000. COCO instance segmentation challenge pre-trained weight was used as the initial weight. We trained the model for 28,000 iterations.

\textbf{Semantic Segmentation:} We trained a ResNet-50 FPN model with two P4 GPUs. The batch size was 16 and the initial learning rate was 0.02. Again, the original high-resolution image with dimensions $1024 \times 1024 \times 3$ was directly used as the training images. ImageNet pre-trained weight was used as the initial weight. This time, we trained the model for 50,000 iterations.

\begin{figure}[h!]
\begin{subfigure}{.9\textwidth}
    \includegraphics[width=\linewidth]{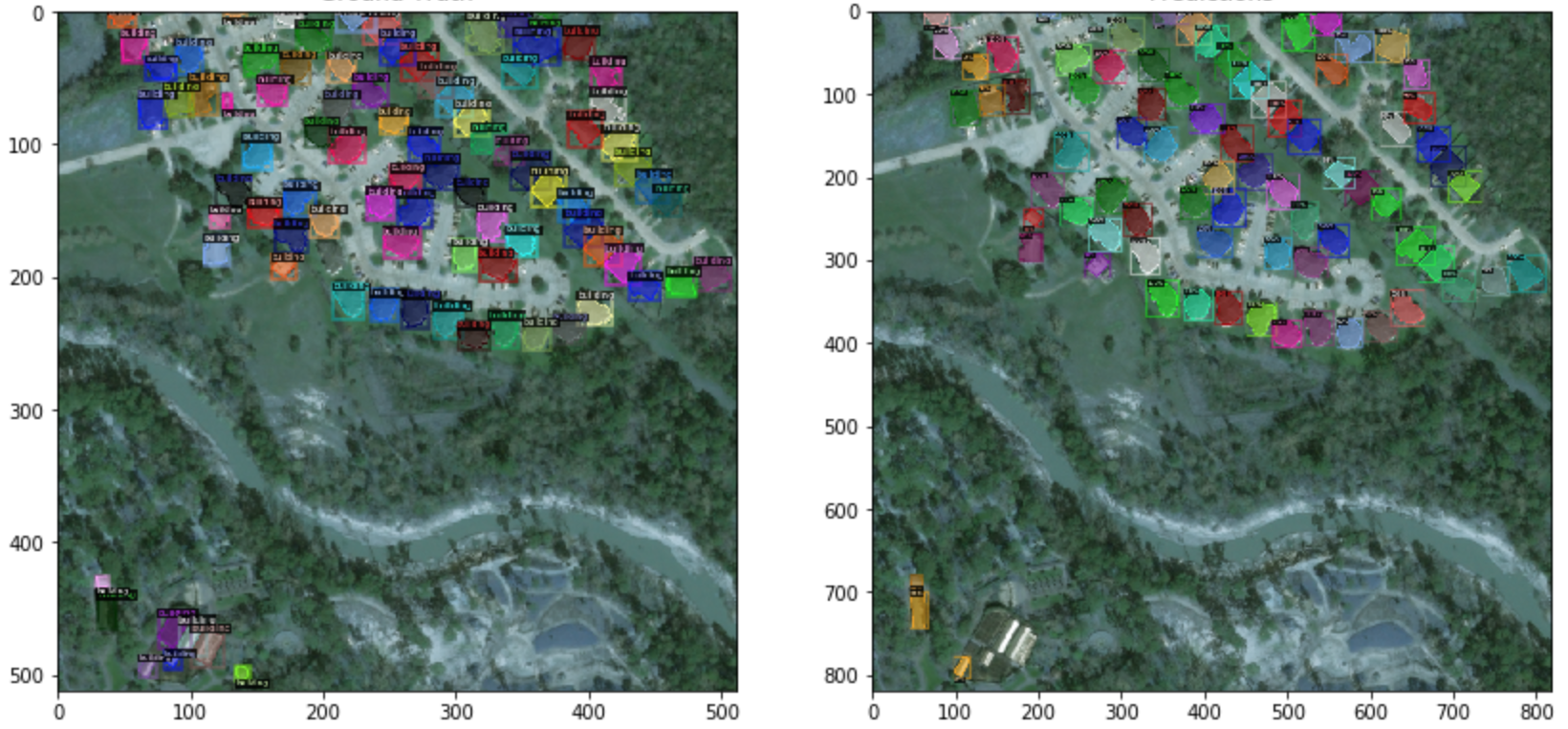}
    \caption{Instance segmentation results on validation set}
    \label{fig:ins_seg}
\end{subfigure}
\newline
\begin{subfigure}{.9\textwidth}
    \includegraphics[width=\linewidth]{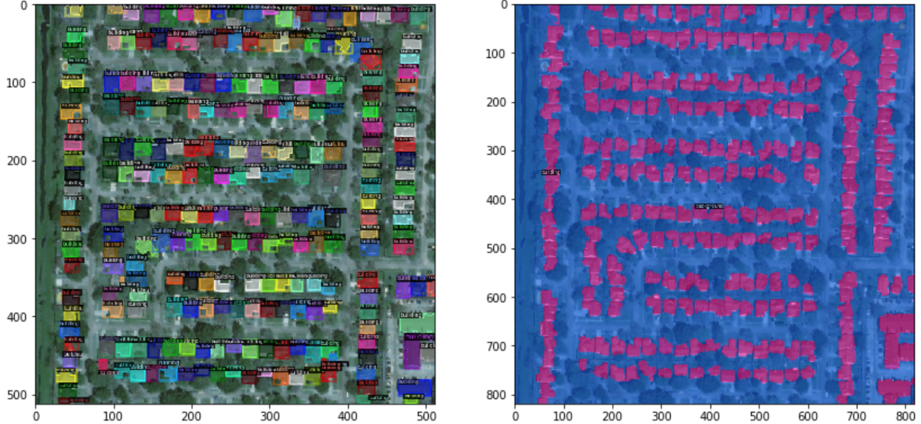}
    \caption{Semantic segmentation results on validation set}
     \label{fig:sem_seg}
\end{subfigure}
   \caption{Building segmentation results on validation set. Left column is the ground truth data. The different colors represent each labeled individual building. a) right is the instance segmentation prediction. The different colors represent each individually detected instance of buildings. b) right is the semantic segmentation prediction. The prediction is a binary mask, with background (blue) and buildings (red) is obtained.}
%\label{fig:classification_model}
\end{figure}

The prediction result of the instance segmentation and semantic segmentation model is shown in Figure~\ref{fig:ins_seg} and Figure~\ref{fig:sem_seg} respectively. To conduct a consistent comparison of the two models, we computed the metric scores only based on their predicted masks. The predicted mask of Instance segmentation are a integration of all the individual masks of each building. We compared the mIoU score of our models on our validation set in Table~\ref{tab:seg_results}. In addition, we compared our results to the xView2 challenge baseline. Though it seems that both models performed relatively well through visual inspection (Figure~\ref{fig:ins_seg} and Figure~\ref{fig:sem_seg}), the results show that our semantic segmentation model performs significantly better than the xView2 challenge baseline as well as our instance segmentation model. 

The inferior performance of the instance segmentation model was surprising. We realized that one of the major issue of using Mask R-CNN for instance segmentation is its inability to detect all the buildings, particularly where there are more than 1000 buildings in a single image. We believe that using a larger and more complex model can improve the performance of the instance segmentation model. However, given the difficulties of optimizing the instance segmentation model as well as the fact that our semantic segmentation model was able to achieve relatively accurate results, we chose the semantic segmentation model for our final model.

\begin{table}
\begin{center}
\begin{tabular}{|l|c|}
\hline
Method & mIoU  \\
\hline\hline
xView2 Baseline & 0.78 \\
Instance Segmentation & 0.70 \\
Semantic Segmentation & 0.85 \\
\hline
\end{tabular}
\end{center}
\caption{Building segmentation results on the validation set. mIoU is calculated based on the predicted binary mask of building footprint}
\label{tab:seg_results}
\end{table}

\subsubsection{Building Classification}
As discussed in Section \ref{classification_model}, we implemented several different architectures for our damage classification models. 

\textbf{Pre \& Post Concatenation:} Inspired by the work done by Xu et al. \cite{xu2019building}, who studied a task highly analogous to the one we perform here, we implemented a twin-tower model which concatenates the output features of a ResNet-50 model from pre- and post-disaster building cropped images.  

% Finally, an additional model was built outside of the DeepLab pipeline. In this model we directly concatenated pre- and post-disaster cropped images of buildings. Using an architecture similar to the end-to-end model that was explained in Section 4.2.1, we used a pretrained Resnet34 and modified the first layer of the Resnet to accommodate the 6 channel input of our concatenated image.

\textbf{Pre \& Post Concatenation with Disaster Labels:} Acting off the knowledge that expected damage level distribution is different depending on the disaster, we added an additional feature as a third branch before concatenating and feeding into the fully connected layers. The additional feature is a label denoting the disaster to which a building belonged (e.g.  ``1" for Hurricane Matthew). Note that these labels are not available in the xView2 test set. To appropriately classify test set images with this model, we built an additional CNN model to identify image disaster name based on pre- and post-disaster imagery. This disaster name classification task was accomplished with this CNN to a high level of accuracy, achieving 95\% accuracy on the validation set. 
% Due to time constraints we were unable to incorporate this model into our pipeline as of publication, so we do not report results on the xView2 challenge test set in this paper.

% \textbf{Pre Minus Post:} \cite{xu2019building}, which studied a task highly analogous to the one we perform here, found that their disaster classification method worked best when they included in their model a direct subtraction of pre-disaster and post-disaster feature maps. Inspired by this analysis, we built a model which took as input the pixel-wise subtraction of cropped building footprints from the pre- and post-disaster images. 

\textbf{Pre \& Post Concantenation with SSIM:}
We experimented with adding hand-engineered features to our models to boost performance. Here we used SSIM, a metric that is commonly used in image processing to measure the similarity between two images. Similar to the model with disaster label, this additional feature is added as a third branch of our architecture before concatenation.

All models were trained using cross-entropy loss for damage classification score. The models were trained using one NVIDIA Tesla T4 GPU with a batch size of 500 and initial learning rate of $8 \times 10^{-4}$ and were trained until model convergence. These models were first tested for appropriate implementation by over-training on a smaller subset of the data.

\begin{table}
\begin{center}
\begin{tabular}{|l|c|c|c|}
\hline
Method  & Train Acc. & Val.Acc. \\
\hline\hline
Pre \& Post Concat.  & 0.77 & 0.80\\
Pre \& Post Concat. + Disaster  & 0.91 & 0.86\\
Pre \& Post Concat. + SSIM & 0.83 & 0.82\\
\hline
\end{tabular}
\end{center}
\caption{Damage classification task results}
\label{tab:classification_results}
\end{table}

The results of our experiments are summarized in Table \ref{tab:classification_results}. Counter-intuitively, Pre \& Post Concatenation model scored higher on the validation set than on the training set. We attribute this to the model simply memorizing the damage distribution of the dataset, rather than learning anything intrinsic about the input image, and then benefiting from a validation set that had a slightly more favorable distribution of building damage scores. As anticipated, the model with the additional feature of disaster labels show a higher accuracy, which proves that providing the model with information about which disaster a given building came from dramatically improves performance. The model with SSIM as additional feature also performs better than the basic Pre \& Post concatenation model, though still lower than the model with the disaster label. This shows that though knowing the disaster improves the performance, adding additional hand-engineered features can also help with performance. 

% Because the trained models struggled to accurately classify building damage extent beyond simply memorizing the distribution of damage levels, we next trained several models on a simpler task: binary classification of damaged or not damaged buildings. For this task we implemented both a Pre Minus Post model and a Pre Minus Post with Labels model. The results are summarized in Table \ref{tab:binary_classification_results}. As anticipated, providing the model with information about which disaster a given building came from dramatically improves performance.

{\renewcommand{\arraystretch}{1.5}%
\begin{table}[t]
  % \caption{Summary of results submitted to the xView2 challenge Leaderboard}
  \label{tab:test_results}
  \centering
  \begin{tabular}{lccc}
    \toprule
    \cmidrule{1-4}
    Method     & Segmentation F1 Score & Classification F1 Score & Overall F1 Score  \\
    \midrule
{xView2 Baseline}             & 0.80482       & 0.06091  & 0.28408    \\
\hline
{End-to-end Model}     & 0.79584   & /    & 0.23875   \\
\hline
Semantic Segmentation +  \\Pre \& Post Concatenation  & 0.84330  & 0.51  & 0.61 \\
\hline
\textbf{Semantic Segmentation} + \\ \textbf{Pre \& Post Concatenation + Disaster} & \textbf{0.84330} & \textbf{0.5873}&\textbf{0.6641}  \\
\hline
 Semantic Segmentation +  \\Pre \& Post Concatenation + SSIM    & 0.84330  & 0.54679 & 0.63574   \\ 
 \bottomrule
  \end{tabular}
\caption{Summary of results submitted to the xView2 challenge Leaderboard}
\end{table}
}

% \begin{table*}[htp]
% \begin{center}
% % \resizebox{\textwidth}{!}{%
% \begin{tabular}{|c|c|c|c|c|}
% \hline
% \multicolumn{2}{|c|}{Method}      & Segmentation F1 Score & Classification F1 Score & Overall F1 Score 
% \\ \hline\hline
% \multicolumn{2}{|c|}{xView2 Baseline}                                                                   
% & 0.80482       & 0.06091  & 0.28408                   
% \\ \hline
% \multicolumn{2}{|c|}{End-to-end Model}     
% & 0.79584   & /    & 0.23875                  
% \\ \hline
% \multirow{3}{*}{Two-Step Model} 
% & Semantic Segmentation +  Pre \& Post Concatenation                    
% & 0.84330                        & 0.51                             & 0.61                      
% \\ \cline{2-5} 
% & \textbf{Semantic Segmentation +  Pre \& Post Concatenation + Disaster} & \textbf{0.84330} & \textbf{0.5873} &\textbf{0.6641}           
% \\ \cline{2-5} 
% & Semantic Segmentation +  Pre \& Post Concatenation + SSIM    & 0.84330  & 0.54679 & 0.63574                   
% \\ \hline
% \end{tabular}%
% % }
% \end{center}
% \caption{Summary of results submitted to the xView2 challenge Leaderboard}
% \label{tab:test_results}
% \end{table*}

\begin{figure}[h!]
\begin{center}
    \includegraphics[width=5in]{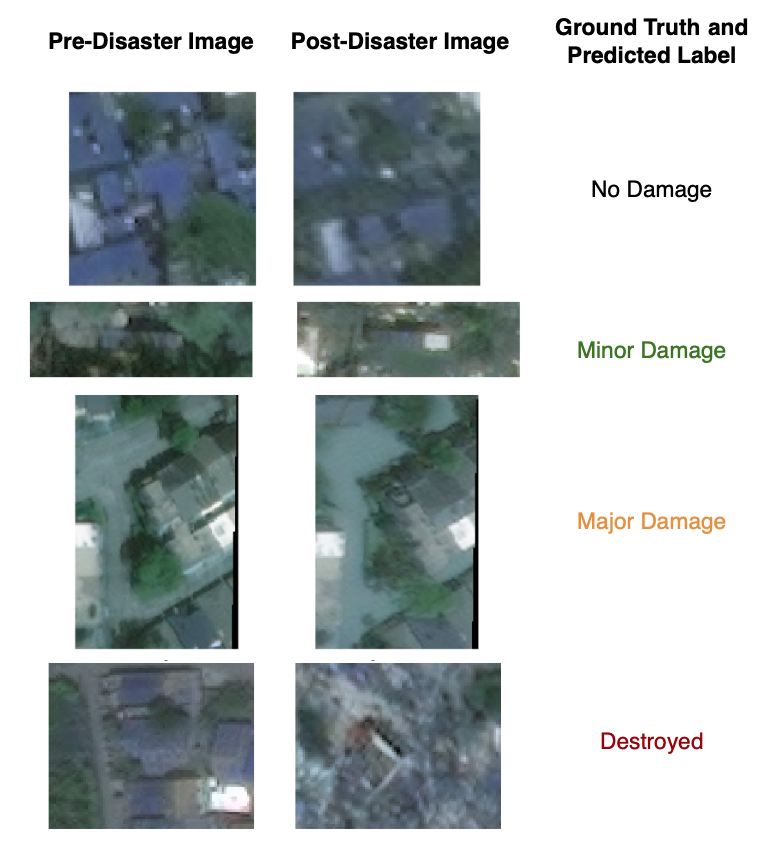}
\end{center}
   \caption{Validation results of correct damage classification predictions}
\label{fig:correct}
\end{figure}

\begin{figure}[h!]
\begin{center}
    \includegraphics[width=5in]{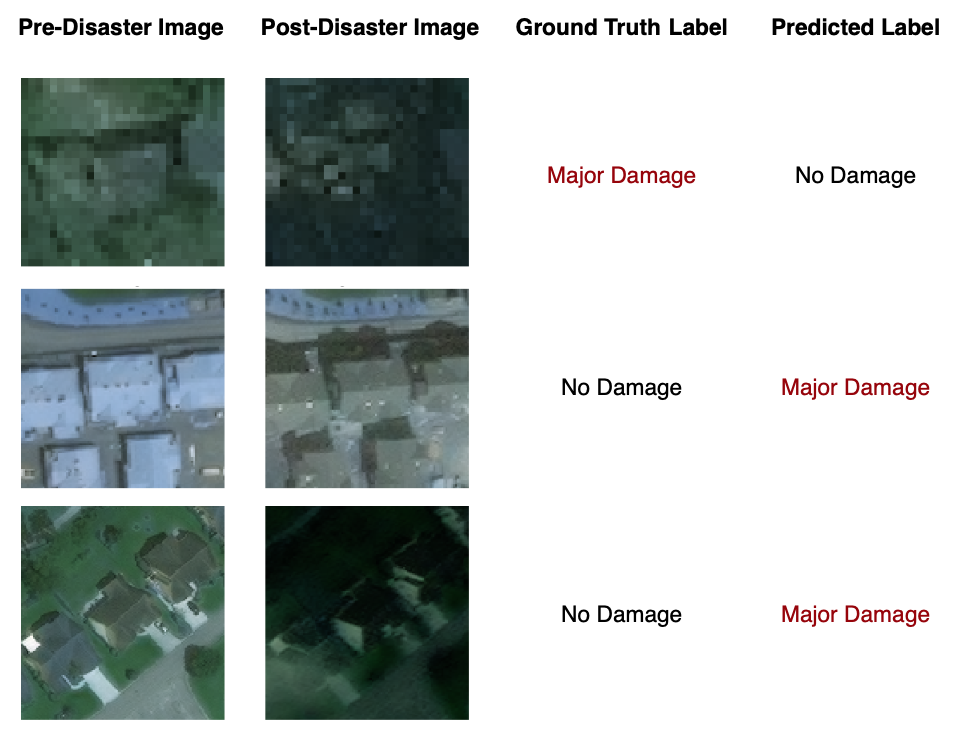}
\end{center}
   \caption{Validation results of incorrect damage classification predictions}
\label{fig:error}
\end{figure}

When visualizing our damage classification predictions, we noticed that our model performs well if both the pre- and post-disaster images are identical in terms of color and nadir angle. This can be seen in Figure \ref{fig:correct}. On the other hand, the model has issues dealing with very small buildings, and when the pairs of images were taken in very different times, causing differences in color, shadows, and nadir angles (Figure \ref{fig:error}). 

\subsection{Test Results}

% \begin{table*}[h!]
% \begin{center}
% \begin{tabular}{|p{0.5\linewidth}|p{0.1\linewidth}|p{0.1\linewidth}|p{0.12\linewidth}|}
% \hline
% Method & Localization F1 Score & Classification F1 Score &  Overall F1 Score \\
% \hline\hline
% xView2 Baseline & 0.80482 & 0.06091 & 0.28408 \\
% End-to-end model & 0.79584 & 0.00000 & 0.23875\\
% Two-step model w/ Pre \& Post Concat. & 0.84330 & 0.24982 & 0.42787 \\
% \textbf{Two-step model w/ Pre \& Post Concat. + Disaster} & \textbf{0.84330} & \textbf{0.58730} & \textbf{0.66410}\\
% Two-step model w/ Pre \& Post Concat. + SSIM & 0.84330 & 0.54679 & 063574\\
% \hline
% \end{tabular}
% \end{center}
% \caption{Summary of results submitted to the xView2 challenge Leaderboard}
% \label{tab:test_results}
% \end{table*}
For final analysis, we performed inference on the xView2 challenge test dataset and submitted the results to the xView2 leaderboard for scoring. These test dataset results were generated for the End-to-End Model and the two step models. For all two-step models, we used the output of the semantic segmentation model, which performed better in our experiments. Because the test set images do not include information about source disaster, we used the separate CNN model discussed earlier to classify test images by disaster name. These names were then included in the model which used disaster name as a feature. Results are summarized in Table \ref{tab:test_results}.

The End-to-End model performed only slightly worse than the xView2 baseline on the segmentation part, a somewhat surprising result given the model's significantly less complex design. The classification f1 score of the End-to-End model is not available, which we attribute to a submission error which could not be resolved by the time of writing. However, due to its relatively poor performance in training, we expect that the End-to-End model would not perform well on the test set. Both two-step models performed better in localization and classification compared to baseline. Our best model, a two-step model with Pre \& Post Concatenation and disaster label, scored significantly better than baseline (overall F1 score of 0.66 versus 0.28 for baseline).

% \begin{table}
% \begin{center}
% \begin{tabular}{|l|c|c|c|}
% \hline
% Method  & Train Accuracy & Val.Accuracy \\
% \hline\hline
% xView2 Baseline  & 0.76 & 0.81\\
% Pre Minus Post  & 0.84 & 0.73\\
% \hline
% \end{tabular}
% \end{center}
% \caption{Damage classification task results}
% \label{tab:classification_results}
% \end{table}

% \begin{table}
% \begin{center}
% \begin{tabular}{|p{0.3\linewidth}|c|c|c|}
% \hline
% Methods  & Train Accuracy & Val.Accuracy \\
% \hline\hline
% Pre Minus Post  & 0.75 & 0.73\\
% Pre Minus Post w/ Labels  & 0.94 & 0.84\\
% \hline
% \end{tabular}
% \end{center}
% \caption{Binary damage classification task results}
% \label{tab:binary_classification_results}
% \end{table}

\section{Conclusion}

In this paper, we presented a framework for building damage assessment after natural disasters. We experimented with several different architectures that focused on building segmentation and classification, in particular an end-to-end model that simultaneously performs those two tasks and a two-step model. Our approach presents one of the first attempts at creating a model that is able to generalize to all types of natural disasters. 

We submitted our results as part of the xView2 Challenge, a competition to design better models for identifying buildings and their damage level after exposure to multiple kinds of natural disasters. Our best model couples a building identification semantic segmentation convolutional neural network (CNN) to a building damage classification neural network, with a combined F1 score of 0.66, surpassing the xView2 challenge baseline F1 score of 0.28. 

Our results point to several important conclusions. We found that two-step models (segmentation and classification) that are trained individually works better than an end-to-end model that performs the two task simultaneously.

Our best performing model included an additional feature representing the disaster name. This result indicates two possibilities: (1) that similar levels of damage are visually different in different disasters, or (2) given the difficulty of the damage classification tasks, a best performing model will rely heavily on a learned prior probability distribution for damage, which will differ by disaster.

Though this is a huge step towards the automation of building damage assessments for various types of natural disasters, it is clear that there is still a lot of work to be done to increase the accuracy in the building damage classification. Though our results suggest that training a model for each individual disaster might boost performance in the xView2 challenge, we realize that this will not be applicable for new disasters and real-world applications. Instead, our results suggest developing methods that incorporate disaster severity, region vulnerability, and other factors in the form of a prior estimate of damage.

% We were able to achieve significantly better results than the challenge baseline for both the building localization and damage classification tasks. Our best localization model (Resnet-50 FPN) outperformed baseline (F1 score of 0.80 for baseline, 0.84 for our model) while our best classification model (Pre Minus Post) outperformed baseline by a large degree (0.06 for baseline, 0.37 for our model).

% Our results point to a path forward for further progress before the competition end date (Dec 31, 2019). We suspect that we can more easily make progress on the damage classification task than on the localization task, where we have achieved high accuracy already. The binary task results indicate that including information about which disaster a given building was exposed to can dramatically improve performance; we expect to have a working model to incorporate predictions of disaster on the test set within a week. We would also like to experiment with incorporating additional contextual information (e.g.  a scaled version of the entire input image), generating hand-designed features (e.g.  textural analysis metrics), and training much deeper networks.

\section{Acknowledgments}
We would like to acknowledge the Stanford CS325B (Data for Sustainable Development) teaching team for their continuous feedback and ideas. In particular, we would like to thank Professor Marshall Burke, Professor David Lobell and Stefano Ermon, as well as Robin Cheong, Matthew Tan and Burak Uzkent.

\bibliographystyle{unsrtnat}
\bibliography{references}  %%% Uncomment this line and comment out the ``thebibliography'' section below to use the external .bib file (using bibtex) .





% \end{thebibliography}

\end{document}